\documentclass[12pt, letterpaper]{article}

\usepackage[utf8]{inputenc}
\usepackage[T1]{fontenc}
\usepackage{mathptmx} 

\usepackage{amsmath}
\numberwithin{equation}{section}
\usepackage{amsthm}
\usepackage{amsxtra}
\usepackage{amsfonts}
\usepackage{amssymb}
\theoremstyle{plain} 

\usepackage{algorithm}
\usepackage{algorithmic}
\usepackage{dsfont}

\usepackage{hyperref}
\usepackage{graphicx}
\usepackage{caption}
\usepackage{subcaption}
\hypersetup{colorlinks,breaklinks,
	     linkcolor=blue,urlcolor=blue,
	     anchorcolor=blue,citecolor=blue}
\usepackage[table]{xcolor}
\usepackage{rotating}

\usepackage[T1]{fontenc}
\usepackage{booktabs}
\usepackage[margin=1in]{geometry}
\usepackage{adjustbox,lipsum}
\usepackage{threeparttable}
\usepackage{lscape}
\usepackage{array}

\providecommand{\keywords}[1]
{
  \small
  \textbf{\textit{Keywords---}} #1
}

\newcommand{\nc}{\normalcolor}

\usepackage[style=apa,backend=biber]{biblatex}
\bibliography{yezziwoodley}

\usepackage{comment}

\graphicspath{{./images/}}

\usepackage{pdfpages} %
\write18{
pdflatex SI
}

\title{The Batch Artifact Scanning Protocol: A new method using computed tomography (CT) to rapidly create three-dimensional models of objects from large collections \textit{en masse}\thanks{{\bf Source Code:} \url{https://github.com/jwcalder/CT-Surfacing}}}
\author{Katrina Yezzi-Woodley\thanks{Department of Anthropology, University of Minnesota,yezz0003@umn.edu (corresponding author)} \and Jeff Calder\thanks{School of Mathematics, University of Minnesota}	\and Mckenzie Sweno\thanks{Department of Anthropology, University of Minnesota} \and Chloe Siewert\footnotemark[4] \and Peter J.~Olver\footnotemark[3]}
\date{}

\begin{document}

\maketitle

\begin{abstract}
\begin{small}
Within anthropology, the use of three-dimensional (3D) imaging has become increasingly common and widespread since it broadens the available avenues for addressing a wide range of key anthropological issues. The ease with which 3D models can be generated and shared has major impact on research, cultural heritage, education, science communication, and public engagement, as well as contributing to the preservation of the physical specimens and archiving collections in widely accessible data bases. Current scanning protocols have the ability to create the required research quality 3D models; however, they tend to be time and labor intensive and not practical when working with large collections. Here we describe a streamlined \textit{Batch Artifact Scanning Protocol} to rapidly create 3D models using a medical CT scanner. While this method can be used on a variety of material types, we have, for specificity, applied our protocol to a large collection of experimentally broken ungulate limb bones. By employing the Batch Artifact Scanning Protocol, we were able to efficiently create 3D models of 2,474 bone fragments at a rate of less than $4$ minutes per specimen.
\end{small}
\end{abstract}

\keywords{computed tomography, scanning, 3D models, bone fragments, automated surfacing}

\section{Introduction}
\label{sec:intro}

The use of 3D imaging within archaeology is surging in popularity because it expands, in astounding ways, the avenues used for addressing anthropological questions \parencite{hirst2018standardisation, remondino20143d, weber2011virtual}. For example, researchers are able to reassemble fragmentary objects, reconstruct missing structures, mitigate taphonomic distortion \parencite[e.g.][]{benazzi2014virtual, delpiano2017contribution, zollikofer2005virtual, zvietcovich2016novel}, advance geometric morphometric research \parencite[e.g.][]{baab2012shape, baab2013homo, white2022geometric, bastir2019workflows}, improve upon the ways in which data are collected, and to extract new types of data that cannot be collected directly from the object \parencite[e.g.]{yezzi2021virtual, o2020computation, schulz2020brief, yezzi2024using}. And, in the case of computed-tomography (CT) it can be leveraged to non-destructively access otherwise inaccesible internal structures (e.g. the neurocranium, endocranium, or pneumitization and sinuses) \parencite{wu2009application, seidler1997comparative, tobias2001re, ponce1999new}, virtually differentiate fossils from adhering matrix or infilled cavities \parencite{zollikofer1998computer, conroy1984noninvasive, brauer2004virtual, tobias2001re}, or even view mummies inside their encasements \parencite{wu2009application, white2018suitability}. 3D models have been used for studies on biomechanics \parencite{weber2011virtual1, strait2009feeding, o2011combining, spoor1994implications} and allometry and ontogeny \parencite{ponce2001neanderthal, penin2002ontogenetic, massey2018pattern}. For objects where 2D sketches are used widely, such as stone tools and pottery, 3D models have been used to create 2D technical drawings in a more time efficient, consistent, and reliable manner \parencite[e.g.][]{barone2018automatic, magnani2014three, horr2009considerations}. 3D models have been used  to refine typologies \parencite{grosman2008application} and analyze reduction and operational sequences \parencite{clarkson2014determining, clarkson2013measuring, hermon2018integrated}. Zooarchaeologists and taphonomists are using 3D models generated via micro-computed tomography (mCT), micro-photogrammetry, structured light scanning, and high power imaging microscopes to study bone surface modifications and surface texture \parencite{arriaza2019geometric, lopez2019applying, martisius2020method, mate2019new, otarola2018differentiating}. These are but a few examples of the ways in which anthropologists are using 3D models in their research. 

3D scanning has had major impacts for cultural heritage and data sharing. Digital models can be shared electronically making them more accessible to researchers across the globe   \parencite{abel2011digital, wrobel2019digital} through platforms such as \href{https://www.morphosource.org/}{MorphoSource}, \href{https://www.virtual-anthropology.com/}{Virtual Anthropology}, \href{https://sketchfab.com/}{Sketchfab}, \href{https://archaeologydataservice.ac.uk/}{Archaeology Data Service}, \href{https://3d.si.edu/}{Smithsonian3D}, \href{https://africanfossils.org/}{AfricanFossils.org}, and \href{https://core.tdar.org/}{tDAR}. \parencite[For further discussion on 3D data repositories, see][]{hassett2018bone, bastir2019workflows, wrobel2019digital, mulligan2022data}. The ability to share digital models is especially pertinent for limiting research-related travel, thereby reducing its environmental impact, as well as maintaining continuity during disruptive events such as a pandemic. The ease with which models can be shared expands the possibilities for cultural heritage, education, science communication, and public engagement. Research quality 3D models can be used to facilitate preservation by limiting the handling of the actual object \parencite{means2013virtual, pletinckx2011virtual}. Furthermore, data collection from 3D models is inherently non-destructive \parencite{wu2009application}. Not only are many repositories open-access resources, public institutions are increasingly creating virtual experiences that allow patrons anywhere in the world to explore archaeological sites and museums such as can be found on the Archaeological Institute of America's \href{https://www.archaeological.org/virtual-education-resources/}{online education resource list}. As a result of the push to create public-facing resources, publications have emerged describing methods for creating virtual exhibits and to explore ways in which 3D scanning can be used to engage the public \parencite[e.g.][]{abel2011digital, bruno20103d, younan2015digital, quattrini2020digital, tucci2011effective}. Additionally, options are becoming available for educators to develop content that is more accessible through the application of 3D printing \parencite{evelyn2020getting, bastir2019workflows, weber2014another}. As the field grows, journals (e.g. \href{https://www.journals.elsevier.com/digital-applications-in-archaeology-and-cultural-heritage}{Digital Applications in Archaeology and Cultural Heritage}, the \href{https://polipapers.upv.es/index.php/var}{Virtual Archaeology Review}, and \href{https://journal.caa-international.org/}{The Journal of Computer Application in Archaeology}), conferences and professional organizations (e.g. \href{https://caa-international.org/}{Computer Applications and Quantitative Methods in Archaeology}) are being established that are specifically devoted to the advancement of digital methods in archaeology. Finally, the rapid growth of the field has inspired conversations on the ethics of and best practices for engaging in digital methods within archaeology \parencite[See][]{dennis2020digital, lewis2019fight, 
richards2017catch, white2018suitability, mulligan2022data} such as the FAIR (Findable, Accessible, Interoperable, and Reusable) guiding principles for scientific data management and stewardship \parencite{wilkinson2016fair} and the CARE (Collective benefit, Authority to control, Responsibility, and Ethics) Principles for Indigenous Data Governance established by the Global Indigenous Data Alliance (\href{https://www.gida-global.org/}{GIDA}). 

Photogrammetry, laser scanning, and structured light scanning are commonly used methods for creating 3D models of objects and are useful for creating high-resolution, textured scans. \parencite[See] [for literature detailing these approaches to scanning.] {lauria2022detailed, linder2016digital, magnani2020digital, niven2009virtual, porter2016simple} Though \textcite{bretzke2012evaluating} demonstrated how two objects can be scanned at a time, generally objects are scanned individually and the scanning and post-processing times can be extensive. On the other hand, multiple objects can be scanned simultaneously using medical or micro-CT. The output Digital Imaging and Communications in Medicine (DICOM) files are then interactively surfaced and then separated into individual files and cleaned using expensive, proprietary  and GUI-based software such as Slicer, Aviso, or Geomagic \parencite{goldner2022practical}. In these cases the scanning process is efficient but at the cost of an increase in labor-intensive post-processing time and, like the other methods, may not be conducive for efficiently creating a full set of 3D models of objects from large collections. 

The ability to feasibly scan large collections like faunal assemblages, which can be comprised of over ten thousand specimens, necessitates a significant decrease in the time required for processing and post-processing. As such, scanning has been mostly restricted to small collections or a small subset of a large collection, which imposes limitations on research requiring larger sample sets. The ability to expediently scan and model specimens from large collections opens possibilities for new types of data accumulation, which in turn provides access to methods such as machine learning and other powerful computational approaches designed to handle larger and richer data sets \parencite{zollikofer1998computer,jordan2015machine,carleo2019machine,yezzi2024using,calder2022use, mcpherron2021machine}. 

In this paper, we describe a method that we have developed, which we call the \emph{Batch Artifact Scanning Protocol}, to safely and rapidly scan large assemblages of objects using a medical computed tomography (CT) scanner and compare our results to published results using other scanning methods, particularly photogrammetry and structured light scanning to illustrate the efficacy and efficiency of our method. While \textcite{goldner2022practical} independently developed a similar packaging method for bladelets scanned via mCT, our method emphasizes automated post-processing workflows and is adaptable for different artifact types and imaging modalities. 

For our research purposes we scanned experimentally broken ungulate limb bones. The DICOM data are automatically segmented and surfaced using an algorithm that can be executed in Python. Here we provide a step-by-step description of how to use the Batch Artifact Scanning Protocol, emphasizing key points necessary for achieving optimal results. We then discuss the factors to consider when weighing options for constructing 3D models for research and highlight when this approach is most useful. The purpose of this paper is to provide sufficient details in order to offer an inroad for those who are new to CT scanning, so that the Batch Artifact Scanning Protocol can be adopted and built upon by independent research groups. 


\section{Materials and Method}


\subsection{Materials}

Our sample was comprised of $2,474$ bone fragments drawn from a collection of experimentally broken ungulate appendicular long bones at the University of Minnesota that are being used in research on how early hominins used bone marrow as a food resource.  

\begin{figure}[ht]
\centering
\captionsetup{margin=2cm, justification=centering}
\includegraphics[width=0.75\textwidth]{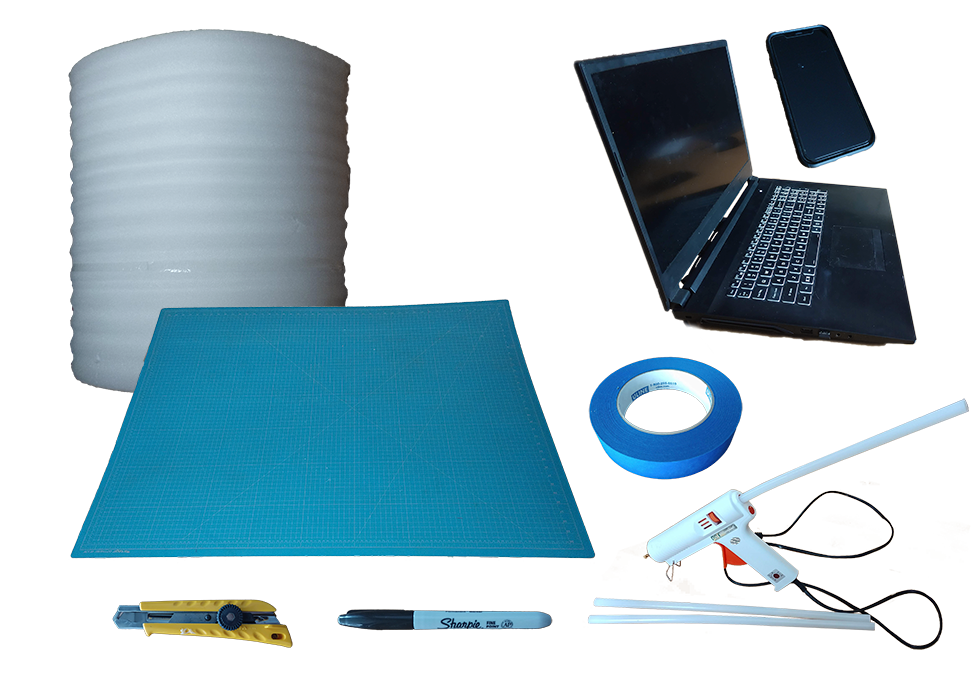}
\caption[Supplies needed for scanning.]
{Supplies for scanning \medskip \par \small Shown here are the supplies we used for scanning which include polyethylene foam, a cutting mat, painter's tape, a hot glue gun with glue sticks, a utility knife, a smart phone for taking photographs for the purpose of documentation, and a laptop to create the \emph{.csv} companion file. While this setup was chosen with the protection of the scanned objects in mind, it should be noted that any packaging material can be used as long as its density is discernibly different from the target object during scanning.}
\label{fig: materials}
\medskip
\end{figure}

We encased bone fragments in foam packets to facilitate the batch scanning process and to offer protection to the specimens during transport to the facility as well as during scanning.  The following materials were needed to create the scan packets: large rolls of polyethylene foam, a glue gun, glue sticks, box cutters, and tape (we used painter's tape). Packets were labeled using a sharpie. The work was done on a large cutting mat to protect the surface of the work table. Large military grade duffel bags were used to transport scan packets to the facility where they were scanned. We also used a computer and camera (or smartphone) to document the fragment layout in a \emph{.csv} file and take photos as backup (See \hyperref[fig: materials]{Figure} \ref{fig: materials}).

\subsection{Methods}
Here we describe in detail the steps and procedures in the Batch Artifact Scanning Protocol, to produce high resolution 3D surface models. This is broadly a three stage process: (1) Preparing specimens for scanning, (2) scanning, and (3) post-processing scans. 

Prior to scanning we assembled the required materials; chose the specimens we wished to scan; and, then bone fragments were placed in scan packets that were then transported to the scanning facility. Each specimen was carefully documented by specimen number in a 
\emph{.csv} file for cross-referencing during post-processing. Fragments were scanned using medical Computed Tomography (CT). Subsequent DICOM data were processed using a Python algorithm that separated each fragment into individual files; segmented the fragment from the remainder of the image (what can be thought of as negative space); and then the segmented images were surfaced to create a 3D mesh. Surfacing is the process by which scan data are converted into a mesh that covers the surface of a solid object, which can then be used for computations and further processing. These meshes were then checked and adjustments were made on individual fragments as needed. The time required for such manual interventions was minimized through our overall efficient and effective design of the scanning and segmentation protocols.

\subsubsection{Preparing specimens for scanning}

\begin{figure}[ht]
\centering
\captionsetup{margin=2cm, justification=centering}
\includegraphics[width=1.00\textwidth]{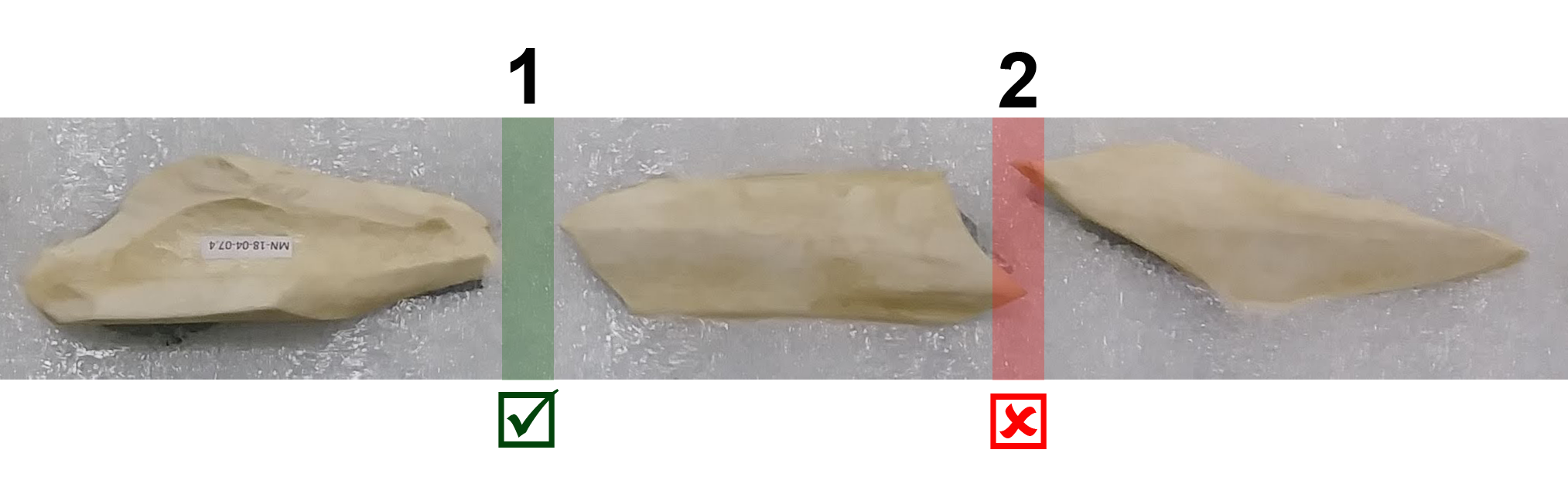}
\caption[Fragment placement]
{Fragment placement \medskip \par \small The fragments should not overlap in the $x-$ or $y-$ directions. This ensures that the automated segmentation can properly separate the fragments within the scan data into individual models for surfacing. The $x-axis$ is the view from the side of the scanning bed. The $y-axis$ is the bird's eye view of the scanning bed.}
\label{fig: spacing}
\medskip
\end{figure}

Bone fragments were packaged in polyethylene foam for transport and scanning. We cut out strips of polyethylene packaging foam and laid the bone fragments linearly, end-to-end along the center of one of the foam strips to provide protection during transport and scanning. In order for the automated segmenting and surfacing algorithm to work properly, we allowed approximately $1$-$2$ cm clearance between bone fragments in each packet. In order for the algorithm to successfully divide the scan into individual fragment files, there can be no overlap between specimens in the $x$- or $y$-directions (see \hyperref[fig: spacing]{Figure} \ref{fig: spacing}). The surfacing algorithm detects breaks between the fragments in the scan data and automatically separates the images according to those breaks. If there is overlap between two fragments the algorithm may not recognize them as two fragments and may combine them into a single fragment or cut off parts of one or both fragments. We also ensured a  $3-5$ cm margin along the edges of the packaging material to accommodate the glue used to seal the packets closed.   

For each packet, we chose fragments that are similar in size in the $x$- and $y$-direction in order to conserve material because several strips of foam were used for each packet. We stacked the foam strips one on top of the other until the stack was high enough to comfortably cover the utmost top edge of the fragment thus providing protection for the specimen in all directions. 

\subsubsection{Documenting Specimens}

\begin{figure}[!t]
\centering
\captionsetup{margin=2cm, justification=centering}
\includegraphics[width=1.0\textwidth]{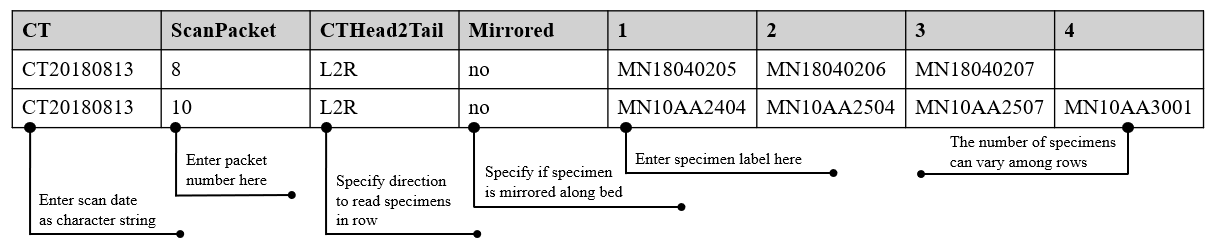}
\caption[Documenting specimens for scanning] 
{Documenting specimens for scanning \medskip \par \small Here we provide an example of how to complete the formatted $.csv$ so that the segmentation and surfacing algorithms will function properly. The first column indicates the date (YYYYMMDD), the second column indicates the packet number for that date, the third column indicates the direction the code should read the \emph{.csv} file, the fourth column indicates whether or not the scan was mirrored, and the remaining columns indicate the specimen labels. The third and fourth columns are there to mitigate the need to resurface the scan should it have been oriented improperly on the scanning bed.}
\label{fig: scanlayoutexample}
\medskip
\end{figure}

Documenting the layout of specimens is a key component of the preparation process, and ultimately post-processing. Once the bone fragments were arranged in the desired order along the first packaging strip, the specimen catalog number was recorded in a \emph{.csv} file (see \hyperref[fig: scanlayoutexample]{Figure} \ref{fig: scanlayoutexample}). (A template can be downloaded from the \href{https://amaaze.umn.edu/software}{AMAAZE website} and \href{https://github.com/jwcalder/AMAAZETools}{GitHub}.) It is essential to adhere to the prescribed \emph{.csv} file format for the algorithm to function properly. It is equally important to label the ``head'' of the packet such that it is placed properly on the CT scanning bed. Each line item in the \emph{.csv} file is dedicated to one packet and the algorithm reads the line from left to right (preferably ``head'' to ``foot'' on the scanning bed). Therefore the leftmost entry in the \emph{.csv} file is generally at the ``head'' of the packet. Should there be an error when the packet is placed on the bed such that it is scanned from ``foot'' to ``head'', then the entry for the column labeled ``CTHead2Tail'' can be changed to R2L so that the line is read in the opposite direction. This happened a couple of times due to user error and our method can easily handle it in post-processing.

\begin{figure}[!t]
\centering
\captionsetup{margin=2cm, justification=centering}
\includegraphics[width=1.0\textwidth]{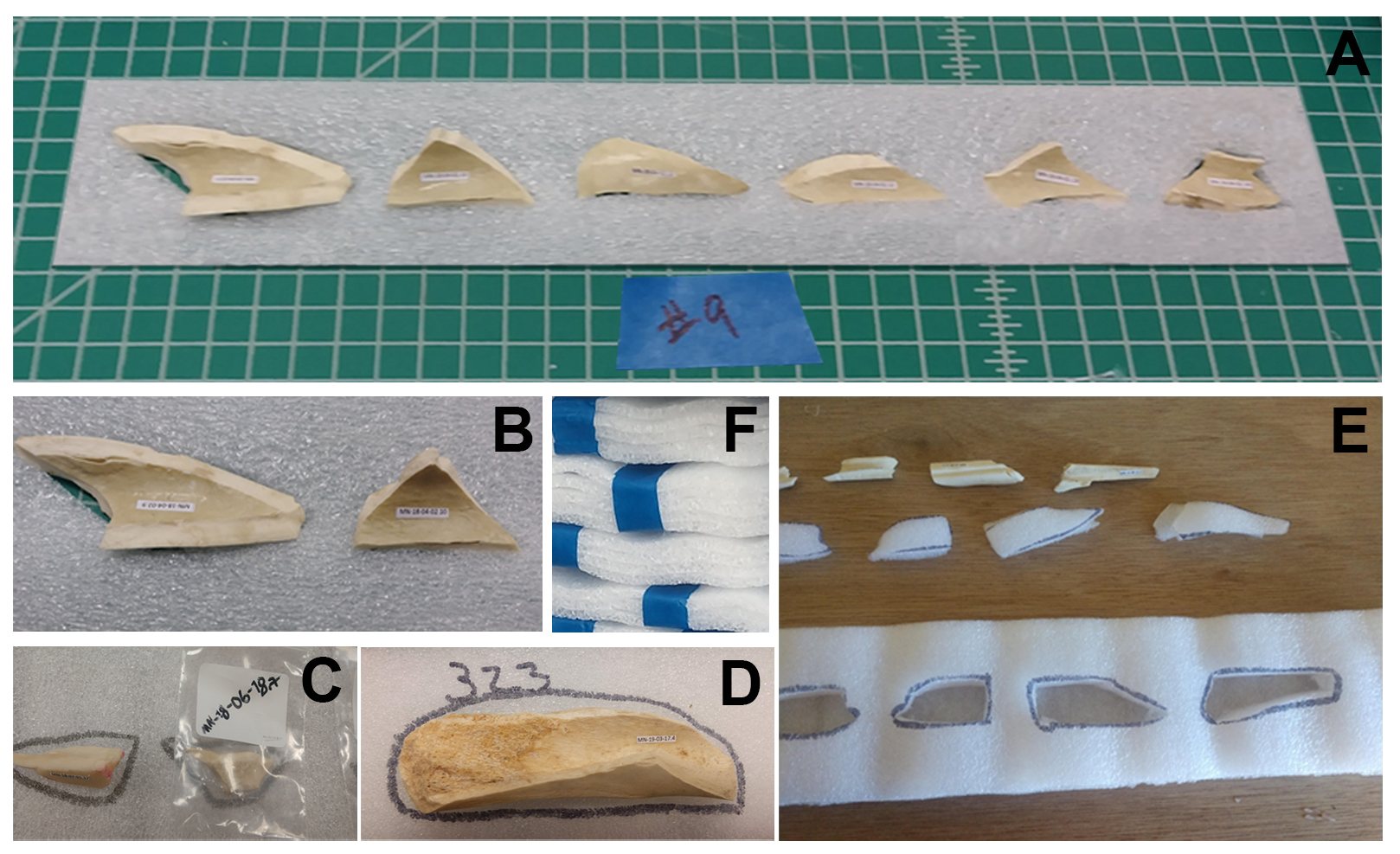}
\caption[Fragment layout]
{Fragment layout \medskip \par \small Here we offer images of the various stages of the packaging process. They are lettered according to the order of operation within the protocol. We took a photograph of the layout of the fragments for the entire package (A). Photographs were taken of individual fragments such that we could clearly see the labels (B). If the fragment was not directly labeled we included the labeled bag in the photograph (C). We traced the fragments using a sharpie (D) and then used the outline to cut out sections in the foam to encase the fragments (E).  Packets were wrapped in tape for additional protection during transport (F).}
\label{fig: process}
\medskip
\end{figure}

We took photographs of how the fragments were laid out on the foam strip for reference and back-up in case there were errors when recording information in the \emph{.csv} file. A wide angled shot of the entire layout of the package was taken, with the package number displayed in the front and center of the image (see \hyperref[fig: process]{Figure} \ref{fig: process}\hyperref[fig: process]{A}). Pictures of the layout of each fragment were taken as well. The specimen number for the individual bone fragment was clearly visible in the image, along with a part of the previous bone specimen for context (see \hyperref[fig: process]{Figure} \ref{fig: process}\hyperref[fig: process]{B}). Some fragments were not directly labeled, but rather stored in individual bags that bore the specimen label. In these instances, the bag was placed above the fragment for the picture (see \hyperref[fig: process]{Figure} \ref{fig: process}\hyperref[fig: process]{C}). Once the packages were sealed, the bags were taped to the outside of the scanning package for ease of repackaging later.

\subsubsection{Packaging Specimens for Transport and Scanning}

Once the layout was established and recorded, we carefully used a sharpie to trace the fragments, being mindful not to get ink on the fragments (see \hyperref[fig: process]{Figure} \ref{fig: process}\hyperref[fig: process]{E}). The outlines were later used as guides for cutting cavities in the foam that add additional protection to the specimens. The outlines can be slightly larger than the fragment itself. A thin tip sharpie can be used for smaller fragments in order to produce a more accurate outline. After outlining, the specimens were removed from the strip and set aside. All but two of the remaining foam strips were glued to the back of the outlined strip using a glue gun. This ensures that the strips will form one cohesive stack (see \hyperref[fig: process]{Figure} \ref{fig: process}\hyperref[fig: process]{F}). Once the glue set, we followed the outlines using box cutters to cut holes in the stack of glued foam. The foam that was removed was set aside in the order that it was cut out so that it could be used later in the process to add padding if additional protection was needed (see \hyperref[fig: process]{Figure} \ref{fig: process}\hyperref[fig: process]{E}). 

After the foam had been cut to create cavities for each bone fragment, the styrofoam stack was flipped upside down to properly attach the bottom piece of the structure. One of the single strips of styrofoam previously set aside was used for the bottom piece. When specimens were heavier, we added additional layers to the bottom and top to prevent the bone fragments from falling out of the packaging during transport and scanning. In some cases, especially when packaging smaller fragments, we cut out the cavities prior to gluing so that we could apply glue near to the edges that were cut to ensure that fragments would not escape the cavity and slip in between the layers of foam.  

Having securely glued the base to the bottom of the package, the stack was flipped right side up and the first specimen cavity (i.e. the ``head'') was placed to the left side to remain consistent with the format in the \emph{.csv} file. We placed individual bone fragments into their corresponding cavities in the same orientation in which they were outlined. Care was taken to place the specimens so they were not likely to move around in transit. As needed, the excess pieces of foam taken from the outline cuttings were used to fill in any gaps between the edge of the cavity and each fragment. This was to prevent unintentional damage related to movement in the cavity and to ensure alignment of the fragments along the z-axis. 

\begin{figure}[ht]
\centering
\captionsetup{margin=2cm, justification=centering}
\includegraphics[width=0.5\textwidth, angle=180]{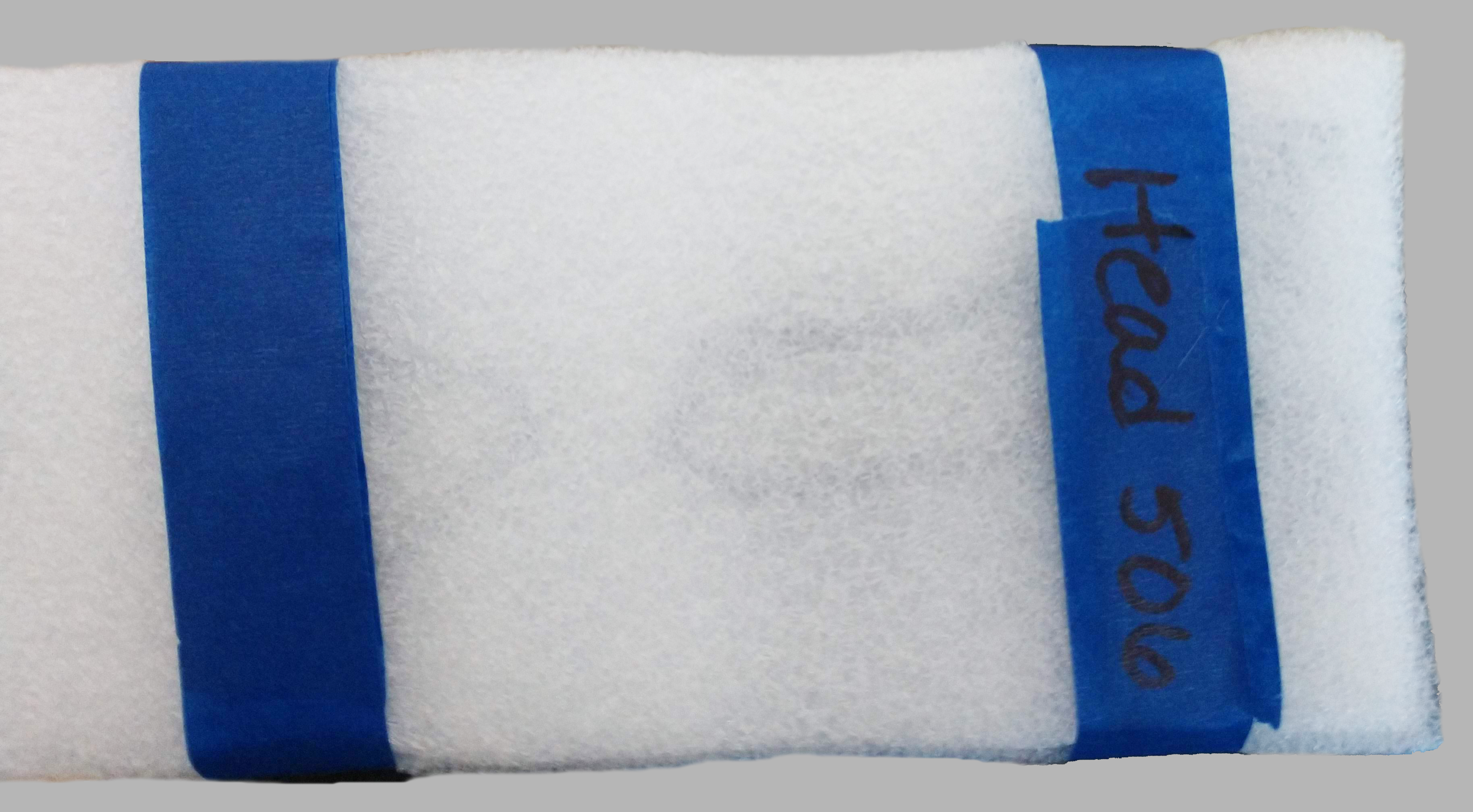}
\caption[Example of package label]
{Example of package label \medskip \par \small Pictured here is an example of how we labeled the package indicating how to orient the package on the scanning bed and as a cross-reference for the \emph{.csv} file for that scan package.}
\label{fig: tape}
\medskip
\end{figure}

\begin{figure}[ht]
\centering
\captionsetup{margin=2cm, justification=centering}
\includegraphics[width=0.5\textwidth, angle =270]{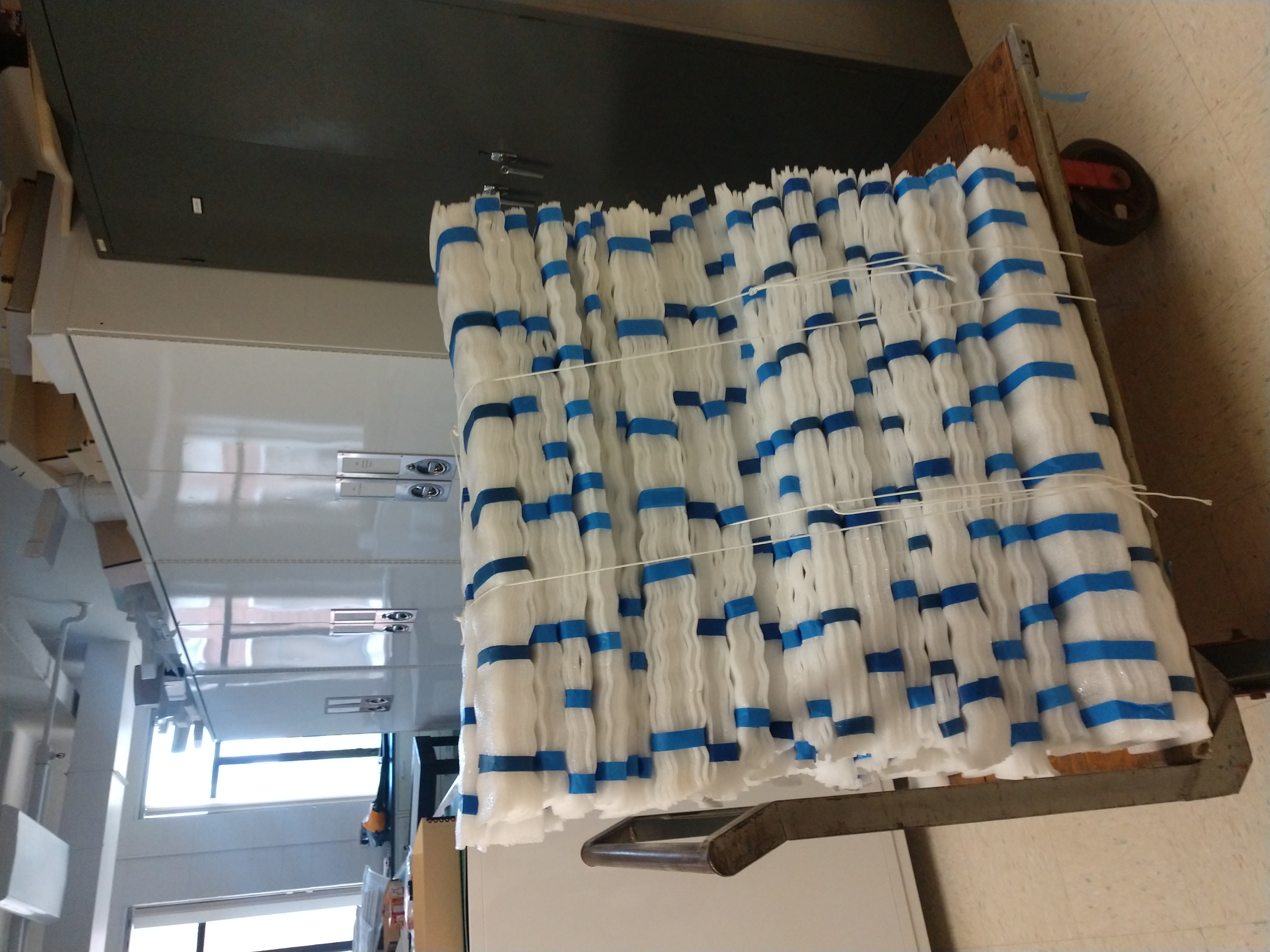}
\caption[Packets ready to go to the scanning facility]
{Packets ready to go to the scanning facility \medskip \par \small Prepared packets are designed for rapid placement onto the scanning bed, minimizing handling time and maximizing efficiency. This approach significantly reduces scanning time and costs, especially in facilities that charge by the hour.}
\label{fig: ready}
\medskip
\end{figure}

Once satisfied with the placement of the bones inside the package and their relative inability to move around in transit, the final foam strip was secured to the top of the package. Care was taken not to get glue on the fragments. After constructing the packet, we wrapped a strip of painter’s tape around the entire shorter circumference at the head of the packet then labeled the tape with the packet number (see \hyperref[fig: tape]{Figure} \ref{fig: tape}) and the word ``head'' so that the CT technicians we worked with knew how to place the packet on the scanning bed and what number to use in the filenames for the output data. Additional strips were added to any section of the packet where extra protection seemed necessary, such as in between two larger fragments or in the middle of long fragments that were at risk of breaking. The completed packets were then ready to go to the scanning facility (see \hyperref[fig: ready]{Figure} \ref{fig: ready}). 

\subsubsection{Scanning} \label{scanning}

We brought a total of $2,474$ bones fragments in $329$ packets to the \href{https://www.cmrr.umn.edu}{Center for Magnetic Resonance Research at the University of Minnesota}. Each packet was scanned individually. Each scan takes a couple minutes including the time it takes to lay the packet on the scan bed, adjust the field of view, and take the scan. Once all scans were completed the data were exported as DICOM files. 

\begin{table}[t!]
\vspace{-3mm}
\caption{CT Parameter Settings}
\vspace{-3mm}
\label{tab: param}   
\vskip 0.15 in
\begin{center}
\begin{small}
\begin{sc}
\begin{threeparttable}
\begin{tabular}{ll}
  \hline
Parameter & Setting \\
  \hline
  slice thickness & 0.6 \\
  reconstruction interval & 0.6 mm \\
  KV & 80 \\
  MA & 28 \\
  rotation time & 0.05 sec \\
  pitch & 0.8 \\
  algorithm & bone window \\ 
  convolution kernel & B60f-sharp \\
   \hline
\end{tabular}
\begin{tablenotes}[para,flushleft]
\footnotesize
Note: Please see supplemental information for more details on these parameters. 
\end{tablenotes}
\end{threeparttable}
\end{sc}
\end{small}
\end{center}
\vskip -0.1in
\end{table}

A Siemens Biograph 64 slice PET/CT was used to scan the packets. The scanning parameters are provided in Table \ref{tab: param}. We provide a high level description of these parameters in the supplemental information. For more detailed, technical descriptions of how CT scanning works, see \textcite{scherf2013computed, spoor2000using, buzug2011computed, withers2021x, sera2021computed}. Using medical CT scanning will likely require working with a trained radiologist who can determine the appropriate settings for achieving an optimal image. Thus, a basic understanding of these parameters can be useful for discussing the needs that are particular to the project with the radiologist, e.g. capturing trabecular bone, working with fossilized material, and navigating matrix in-fill or adhesion, all of which can be effectively mitigated using CT. 

\subsubsection{Post Processing}

Detailed written instructions for creating the 3D models from the DICOM scan data can be found on the \href{https://github.com/jwcalder/CT-Surfacing}{AMAAZE GitHub} along with the AMAAZEtools package required to run the CT-Surfacing scripts in Python. Installation instructions accompany the packages on the AMAAZE GitHub. We have also provided data from two scans (scans 8 and 10), each containing four fragments, and an accompanying \emph{.csv} file so users can try the process without having their own DICOM data. In summary, to begin the process of surfacing the scans to create the models, the AMAAZEtools packages must be installed and one must have the appropriate DICOM files and the properly formatted \emph{.csv} file. 

\begin{figure}[ht] 
    \centering
    \captionsetup{margin=2cm, justification=centering}
    \begin{subfigure}[b]{0.40\textwidth} 
        \centering
        \includegraphics[width=\textwidth]{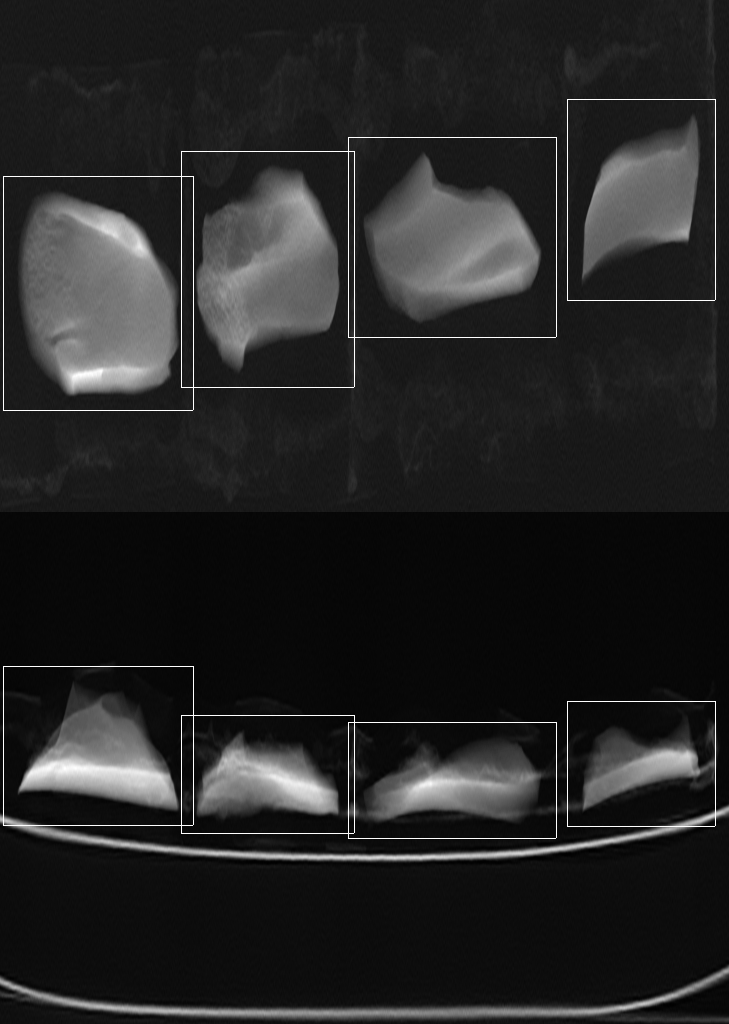}
        \caption*{\textbf{Scan 8}}
        \label{fig: scan8}
    \end{subfigure}  
    \begin{subfigure}[b]{0.402\textwidth} 
        \centering
        \includegraphics[width=\textwidth]{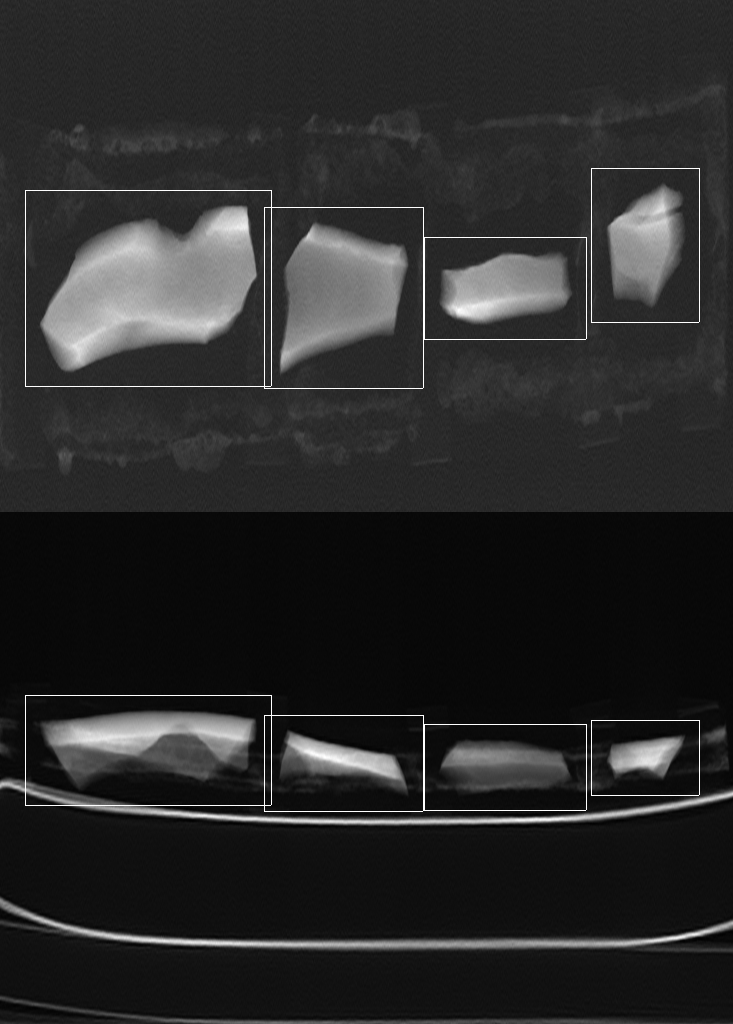}
        \caption*{\textbf{Scan 10}}
        \label{fig: scan10}
    \end{subfigure} 
        \caption[Automated segmentation]
        {Automated segmentation \medskip \par \small Here we provide examples of the .jpg images output by the automated segmentation algorithm which illustrates how the algorithm separates individual fragments from the scan data. These examples come from a session during which we scanned multiple packets. These scans come from packets $8$ and $10$ in that series. Note: These are the scans that are provided along with the source code, offering researchers the opportunity to practice this protocol prior to acquiring their own scan data.}   
        \label{fig: scan8and10}
        \medskip
\end{figure}

The first step is to separate the multi-fragment files into single-fragment files. This is done by running the Python script called \verb|dicom_firstpass.py|, which automatically segments the file into individual fragments and outputs images of the segmentation with bounding boxes (see \hyperref[fig: scan8and10]{Figure} \ref{fig: scan8and10}), as well as the bounding box coordinates, as a \emph{.csv} file which can be manually edited as needed. 

The algorithm for automatically separating multiple bone fragments from a single CT scan works by first thresholding the CT image at a user-specified value in Hounsfield units (HU). The specific threshold depends on the material under consideration; for bone fragments we use 2000 HU. The thresholded binary images are then projected onto each 2-dimensional view of the length of the scanning bed, and the bone bounding boxes are identified by taking the largest connected components of the projected binary images and adding padding on each side. See \hyperref[fig: scan8and10]{Figure} \ref{fig: scan8and10} for a depiction of the computed bounding boxes for each bone for the test scan provided in GitHub.

The automatic algorithm works very well, but there can be occasions when the automatically detected bounding boxes are incorrect. The user can determine this by examining the scan overview images, see Figure \ref{fig: scan8and10}, which depicts the bounding boxes over a two dimensional projection of the scanning bed. In this case, adjustments can be made by editing the \verb|ChopLocations.csv| file that was automatically generated by \verb|dicom_firstpass.py| prior to the next step in the processing. The \verb|ChopLocations.csv| file contains the $(x,y)$ pixels coordinates indicating where the bounding boxes start and stop. The file is initially generated by the automatic algorithm, but can be easily adjusted by the user as needed.   After modifying the \verb|ChopLocations.csv| file, the script \verb|dicom_refine.py| will generate new bounding boxes based on the modified data in the chop locations file, which will also generate new scan overview images, which the user can view to see if the bounding boxes are correct. This process can be iterated several times, if necessary, until the bounding boxes are correctly specified. Again, let us emphasize that the failure cases in the method are very rare and, for the vast majority of the scans, no refinements are needed, although our method makes it easy to refine the bounding boxes when needed. 

Once the files are segmented properly, the next step is to run \verb|surface.py| to generate triangulated surface 3D models for each object in the CT scan. The surfaces are generated from the CT images with the Marching Cubes algorithm; see \cite{lorensen1987marching}. The user must provide a threshold parameter (called \verb|iso| in the code) for the surfacing. As in the segmentation part above, the \verb|iso| threshold value is material dependent, and may also depend on the size of the objects, amount of fine detail, and the resolution of the CT scanner. For surfacing bone fragments we normally use \verb|iso=2500|, with some manual adjustments in special cases.  The surfacing script reads the CT resolutions, which are often different between slices, compared to within each 2D slice, from the DICOM header files and scales the resulting mesh so that the units are millimeters in all coordinate directions.
Choosing good thresholds is largely application dependent. Larger threshold values may omit fine scale details, while small threshold values will pick up on noise and scanning artifacts. When scanning bone fragments, lower values are useful when the bone is extremely thin or more porous and lower values capture trabecular bone better than higher values. 

The \verb|surface.py| script also has the capability to generate rotating animations (as \verb|gif| files) of each object that has been surfaced. All the meshes are output as individual \verb|.ply| files. 

\begin{figure}[ht] 
    \centering
    \captionsetup{margin=2cm, justification=centering}
    \begin{subfigure}[b]{0.95\textwidth} 
        \centering
        \includegraphics[width=\textwidth]{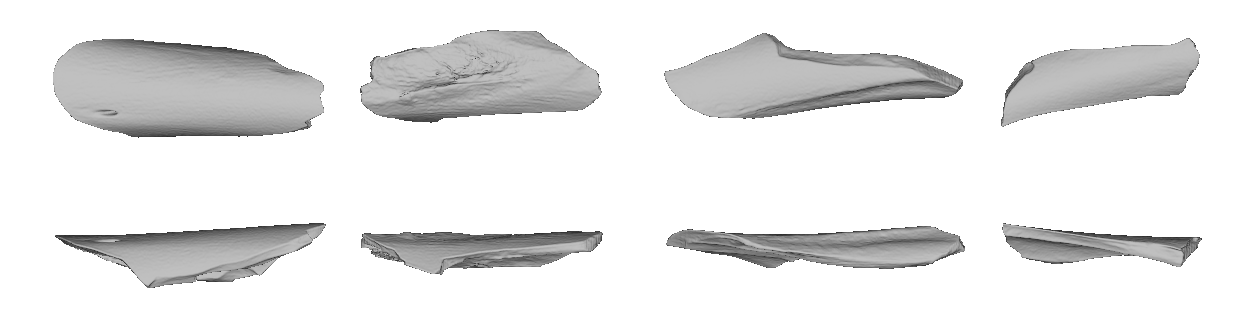}
        \caption*{\textbf{Scan 8}}
        \label{fig: mesh8}
    \end{subfigure}  
    \begin{subfigure}[b]{0.95\textwidth} 
        \centering
        \includegraphics[width=\textwidth]{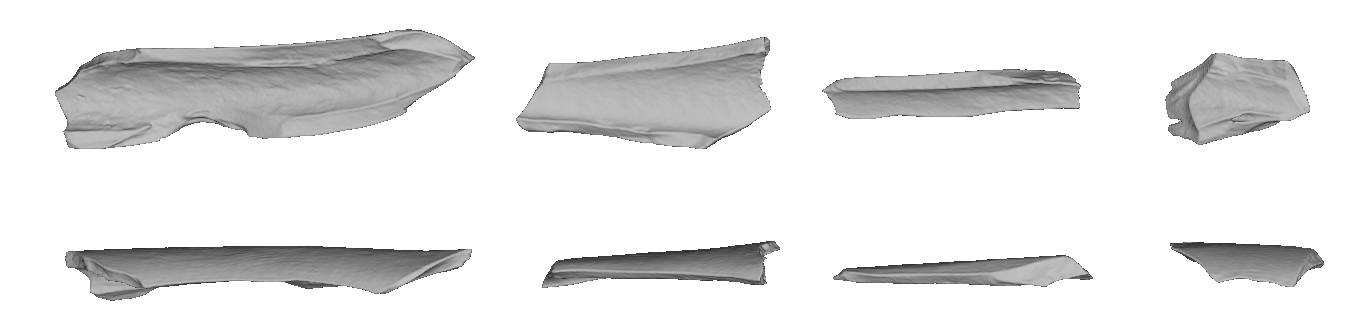}
        \caption*{\textbf{Scan 10}}
        \label{fig: mesh10}
    \end{subfigure} 
        \caption[The final meshes for scans 8 and 10]
        {The final meshes for scans 8 and 10 \medskip \par \small Pictured here are the final meshes from the previous figure of scans 8 and 10.}   
        \label{fig: meshes8and10}
        \medskip
\end{figure}


\section{Results}

Here we compare the Batch Scanning Protocol presented here with other scanning methods and compare our results to scanning and post-processing times that have been published by independent research teams. We attempted other approaches to scanning before deciding to develop the Batch Scanning Protocol and share our experience exploring these options as well as data extracted from the literature. Advancements in all these scanning approaches are ongoing and thus the published literature likely does not reflect the substantial increases in speed of scanning and post-processing that have been achieved to date. Nonetheless, the protocol we present here is a tremendous advancement in this regard and likely surpasses the speed increases not otherwise found in the literature.

For our purposes, photogrammetry proved to be ineffective due to issues of translucency and reflectivity. However, we scanned fragments using the David structured light scanner and David software on a high-end desktop (Dell computer with Microsoft Windows 10 Enterprise OS, 2.7  Quad-Core Intel Core i7, 128 GB RAM). Images were captured every $15^{\circ}$ with a total of 24 captures per $360^{\circ}$ rotation. We found that it takes approximately $5-10$ minutes to set up the specimen. Scanning takes between $15 - 25$ minutes. Typically, it takes at least two rounds of scanning to capture the entire fragment because the fragment needs to be flipped after the first round to capture the portion that was mounted and inaccessible to the scanner in the first round. Once scanning is complete, post-processing is required. For this we used the David software in conjunction with Geomagic Design X, which is a CAD (computer aided design) software package. The time it takes to post process can vary considerably based on the degree to which the scans need to be cleaned and the number of challenges that arise during alignment and registration \parencite[see][]{bernardini20023d}. Based on our experience, it takes on average between $15 - 60$ minutes to post-process the scans to create the 3D model. To set-up, scan and post-process $2,474$ fragments to make 3D models would minimally take $3,505$ hours of interactive user time, or $85$ minutes per fragment.  

On the other hand, with the CT scanning method described here, each packet takes between $15 - 30$ minutes to assemble. We ended up with a total of $329$ packets, and, on average, there were 7.5 fragments per packet. Overall scanning time was $10.75$ hours. The post processing is very fast; it takes 35 seconds to surface all 8 bone fragments in the example GitHub repository using a standard laptop computer, amounting to about 4.375 seconds per fragment. To surface the whole collection takes slightly under 3 hours. This means that to create a single 3D model of a bone fragment using the structured light scanner took $~85$ minutes and $~3$ minutes using medical CT (see \hyperref[tab:time]{Table} \ref{tab:time}).

We compared our time costs to several other examples in recent literature. Unfortunately, in all cases, details of the time costs are missing and thus, for some comparisons, the overall time expense is lower than the likely overall cost. Nonetheless, it is a useful tool for general comparison. \textcite{ahmed2014sustainable} explored the effectiveness and efficiency of an assembly line approach to 3D scanning of artifacts with the help of nine 3D specialists, one of whom scanned and the remaining processed scans. They scanned a total of 300 artifacts comprised of a variety of materials and found the process took roughly 1,260 hours. They used four structured light scanners each set up for different sized objects. Captures were taken every 30 degrees (12 scans) and then flipped and scanned again to capture the other side. Generally 48 scans were required for a complete model and it took about 20 minutes to capture the 48 scans. The macro-scanner and the larger objects (Scanner A) generally took more scans, sometimes up to 96 scans. Since they did not offer a complete breakdown of time costs, we divided the total pre- and post-processing time (1,260 hours) by the number of artifacts (N = 300). Thus, to scan and create a 3D model of one object took 4.2 hours. To scan our sample would take 10,391 hours. It should be noted that this post-processing also involved creating digital museum environments for the display of the 3D models, thus, in this case, the times are likely inflated.

\begin{landscape}
\begin{table*}[ht]
    \caption{Comparison of scanning and processing times}
    \label{tab:time}
    \vskip 0.15in
    \centering
    \begin{threeparttable}
        \begin{tabular}{
            p{0.2\textwidth}
            >{\centering}p{0.065\textwidth}
            >{\centering}p{0.065\textwidth}
            >{\centering}p{0.08\textwidth}
            >{\centering}p{0.085\textwidth}
            >{\centering}p{0.08\textwidth}
            >{\centering}p{0.08\textwidth}
            >{\centering}p{0.085\textwidth}
            >{\centering}p{0.08\textwidth}
            >{\centering}p{0.08\textwidth}
            >{\centering}p{0.08\textwidth}
            >{\centering\arraybackslash}p{0.08\textwidth}
        }
            \toprule
            & BASP & David & Ahmed et al. & Bretzke \& Conard & \multicolumn{2}{c}{Magnani} & Goldner et al. & Porter et al. & \multicolumn{3}{c}{Kingsland} \\ 
            &  &  & (2014) & (2012) & \multicolumn{2}{c}{(2014)} & (2021) & (2016) & \multicolumn{3}{c}{(2020)} \\ 
            & CT & SL & SL & SL & LS & PG & mCT & PG & PG-MS & PG-CC & PG-RC \\ 
            \midrule
            \textbf{Per Specimen:} &&&&&&&&&&& \\
            Set-up (min) & 3.26 & 7.5 & \cellcolor{gray!50}\textbf{NA}\textsuperscript{1} & \cellcolor{gray!50}\textbf{NA} & \cellcolor{gray!50}\textbf{NA} & \cellcolor{gray!50}\textbf{NA} & 0.55 & \cellcolor{gray!50}\textbf{NA} & \cellcolor{gray!50}\textbf{NA} & \cellcolor{gray!50}\textbf{NA} & \cellcolor{gray!50}\textbf{NA} \\
            Scanning (min) & 0.26 & 40 & NA & 9 & 60 & 10 & 0.55 & 12 & \cellcolor{gray!50}\textbf{NA} & \cellcolor{gray!50}\textbf{NA} & \cellcolor{gray!50}\textbf{NA} \\
            Processing (min) & 0.07 & 38 & \cellcolor{gray!50}\textbf{NA} & \cellcolor{gray!50}\textbf{NA} & 40 & 30 & \cellcolor{gray!50}\textbf{1.55} & 88 & 133 & 84 & 63 \\
            Total (min) & 3.59 & 85 & 252 & 9 & 100 & 40 & 2.64 & 100 & 133 & 84 & 63 \\
            \midrule
            \textbf{Overall Sample} &&&&&&&&&&& \\
            \textbf{(N=2,474):} &&&&&&&&&&& \\
            Total (hours) & 148 & 3505 & 10391 & 371 & 4123 & 1649 & 109 & 4123 & 5484 & 3464 & 2598 \\
            \bottomrule
        \end{tabular}
        \begin{tablenotes}[para,flushleft]
            \footnotesize
            \item [] \\
            \item [] \textbf{Table labels and Abbreviations:} BASP=Batch Artifact Scanning Protocol, CT=medical CT, mCT=micro-CT, SL=structured light, LS=Laser Scanner, PG=photogrammetry, MS=MetaShape, CC=ContextCapture, RC=RealityCapture. "Set-up" refers to the time, in minutes, to set-up an individual specimen for scanning. "Scanning" refers to the time, in minutes, to scan an individual specimen. "Processing" refers to the time, in minutes to process the scan data to achieve 3D models. "Total" refers to the time, in minutes to model a single specimen. "Overall" refers to the overall time it would take in hours to scan our sample N=2474.\\
            \item [] \\
            \item [] \textbf{Note:} We reviewed several examples from the literature since 2012 to assess the relative efficiency of our method. In this table, we calculated overall time based on our sample size (N = 2,474). For studies reporting time ranges, we recorded the average value. These examples represent published papers detailing scanning and post-processing times. While we acknowledge, through personal communications and our ongoing work within the realm of 3D modeling, that scanning equipment and techniques are becoming increasingly faster, this progress is not yet fully reflected in the literature. To our knowledge, the protocol we present here remains unmatched by these advancements.\\
            \item [] \\
            \item [1] Some of the data from all comparative samples, such as set-up time and post-processing times, are either missing or partially reported thus the overall time costs for the method are less than what is reported here, perhaps significantly so. Cells with missing data have been highlighted in gray.\\              
        \end{tablenotes}
    \end{threeparttable}
\end{table*}
\end{landscape}

\textcite{bretzke2012evaluating} presented a method to assess variability in the morphology of lithic artifacts (cores and blades) using 3D models thus demonstrating the potential for using 3D models for lithic analysis. They used SmartSCAN by Breukmann to scan and then OPTOCAD software to turn point clouds into a meshes. They scanned cores one at a time and blades two at a time. They said they could scan 5 cores or 10 blades in approximately 1 hour. Their post-processing and set-up times were unspecified so their overall time costs are likely grossly underestimated. That said, a sample might generally have more blades than cores which would offset the additional time costs not reported here. With the available information, it takes 9 minutes to create a 3D model of an individual object and would take 371 hours to scan a collection of 2,474 fragments.

\textcite{magnani2014three} sought to demonstrate the capabilities of laser scanning, and potentially photogrammetry, as methods for replacing hand-drawn lithic illustrations. They scanned various lithic artifacts (Mousterian type scrapers and denticulates, 2 cores, and 1 Achulean handaxe). They used a NextEngine desktop laser scanner with ScanStudio software and they also used photogrammetry with a Canon DSLR and Agisoft software. Using the NextEngine, they took 9 captures (every 40 degrees) per 360 degree rotation. They scanned the object, flipped it upside down and scanned it again. Each orientation took 30 minutes to scan, thus 60 minutes per object. Post-processing took approximately 40 minutes per object. When using photogrammetry, they took about 30 photos per object and it took 10 minutes to photograph each object. The object was stationary and the camera moved around it. Post-processing for a mid-range quality object in Agisoft took 30 minutes per object. They said that high-quality models could sometimes take several hours per object so they chose the mid-range quality. They did not report set-up times in either case, however, we imagine it would be similar to the times we report here when using the David scanner. Regardless, even without the set-up times, according to these data, laser scanning would take 100 minutes per object and 4,123 hours to scan 2,474 objects and photogrammetry would take 40 minutes and 1,649 hours respectively. 

\cite{goldner2022practical} independently developed a batch scanning process they refer to as the StyroStone method. This is the closest to our method as the preparation for scanning is quite similar, however, the post-processing is done manually via Graphical User Interface (GUI) and they scanned bladelets using micro-CT whereas our method focused on medical CT. They then post-processed them in the Aviso and Artec Studio software packages. The authors acknowledge this stating, ``Although the scanning procedure could be accomplished over a short period of time, the subsequent extraction was time consuming.'' (p.~4). Each of their packets could contain up to 220 bladelets. Their post-processing comprised of the following parts: (1) Make the packets, (2) Scan, (3) Model extraction (4) Cropping extracted surfaces (5) Additional cropping, (6) Final cropping. Part 1 takes 2 hours, Part 2 takes 2 hours, Part 3 takes 2 hours for the packet plus an additional minute per item, so in the case of their largest packet 340 minutes. It is not clear how long the remaining multi-step parts (parts 4--7) take. Given that this is a GUI approach it is entirely possible that these parts could take a considerable amount of time. In fact, their time costs (without parts 4--7) are 2.64 minutes per specimen, which would be approximately 109 hours for 2,474 specimens, whereas the method we share here takes 3.53 minutes per specimen and 146 hours overall. The time difference per specimen is 53.4 seconds. It is unlikely that parts 4--7 can be completed in that time frame. Thus, it is likely that our method is still faster even using packets with considerably fewer specimens.

Furthermore, a direct comparison is complicated by the fact that they used micro-CT, which has considerably longer scanning times, whereas we used medical CT. As a result their overall setup and scanning took considerably longer than our scanning time, however, each of their scans contained up to 220 objects whereas our only contained 7 objects on average (see SI for a list of packets and the number of specimens in each packet). We could scale up the number of objects per packet, thus reducing the per specimen scanning time, and our post-processing times would remain exponentially faster. Furthermore, our post-processing method is automated and does not require constant user interface and oversight. If our method were to be scaled up to packages with 220 fragments, we expect the overall time to be approximately 26 hours for a sample of 2,474. 

\textcite{porter2016portable} designed a low-cost portable photogrammetry rig.  It took them 12 minutes to photograph both sides of an object. They did not track their post-processing times for this sample and acknowledge that these times can vary tremendously. However, they did report post-processing times for another ongoing study where they found it took between 43 minutes and 2 hrs and 13 minutes (133 minutes), depending on the desired quality of the model (medium or high). Note that some of the post-processing time is automated.  

\textcite{kingsland2020comparative} focused solely on comparing three different post-processing softwares used in photogrammetry: Agisoft Metashape (MS), Bentley ContextCapture (CC), and RealityCapture (RC). They scanned an aryballos (small, globular flask, 4 cm wide, 18 cm tall) from the Farid Karam Collection. The object was oriented twice (upright and upside down) and 24 images were captured for each of three angles (high angle, middle angle, low angle) per orientation. They calculated average post-processing times and found that Metashape required 133 minutes per object, whereas ContextCapture took 84 minutes, and RealityCapture necessitated 63 minutes.

\section{Discussion}

Choosing an appropriate scanning method for research requires, at minimum, a consideration of the following: (1) portability if required; (2) cost; (3) time; (4) computational resources, including memory, speed, graphics processing units (GPU), and storage; (5) whether or not texture is needed; (6) required scan resolution; and (7) the geometry that needs to be captured.  

The location of the collection that needs to be scanned is the first concern. If the collection cannot be transported to a scanning facility then the scanning equipment must be transported to the collection which eliminates the opportunity for CT scanning. In these cases, the time to scan can increase considerably especially when working with large collections. 

Primary considerations when engaging in 3D scanning are how much time and how much money it will take. Methods like photogrammetry are extremely cost effective \parencite{porter2016simple}. Photogrammetry, laser scanners and structured light scanners are portable and can be taken into the field. However, they are limited to scanning a single object at a time; moreover, it takes a considerable amount of time per scan as compared to the method presented here. Although this is not so problematic when the sample size is small, zooarchaeological assemblages can contain more than $10,000$ specimens, thus making the use of portable scanners untenable. The method presented here thus fills a niche where large quantities of research-quality models need to be created from specimens that can be transported to a CT scanning facility.  

Very few medical CT and micro-CT scanners are portable and it was too cost prohibitive to purchase the equipment ($\ge 100,000$ ~USD) ourselves. Therefore, to use the Batch Artifact Scanning Protocol, we made arrangements to transport specimens to medical scanning facilities and paid fees for scanning services. In total, $329$ packets were required to transport and scan $2,474$ bone fragments. Some facilities will charge per scan and others will charge an hourly rate and rates can vary considerably among institutions and departments. By choosing an hourly rate we were able to scan $2,474$ specimens in $93.75$ hours for $2900$ ~USD.

\subsection{Selecting Specimens to Scan}

When choosing specimens for scanning, one of the most important factors to consider is scan resolution. Choice of the appropriate scan resolution will depend on the research question, the size of the object that needs to be scanned, the capabilities of the scanner, and the field of view used during scanning. Though it may seem counter-intuitive, the highest possible resolution is not always the best option. Optimal resolution depends on the scale required to address the question. Though high resolution scans can serve as a reference, and can always be automatically decimated (i.e., the number of polygons in the surface mesh can be reduced) as required in order to streamline computational algorithms, they can radically increase computer processing times at all stages of the project, including the initial scanning, which may be prohibitive when working with large samples, especially when the size of the files impacts the ability of the computer to store them in memory.

Running test scans is recommended to verify the minimum size required for achieving usable models. For our research purposes, we needed to capture the macromorphology of the bone fragment in order to extract global features and measure angles along the fractured edges of the fragment. We chose to scan fragments that are $\ge$ $2$ cm in maximum dimension but we found that fragments $\ge$ $5$ cm offer more visually appealing models. It is crucial to distinguish between visual appeal and data robustness. A model that looks visually impressive may not be necessary for research purposes, especially if the required data or measurements are more global, i.e. we want to avoid confusing aesthetic detail with research readiness and to assess needs based on the types of data that are to be collected or the purpose of the 3D model (e.g. museum displays vs. research). Other considerations include how thin the bone is (i.e. areas of translucence) and the relative dimensions of the fragment. 

\begin{figure}[ht]
\centering
\includegraphics[width=0.75\textwidth]{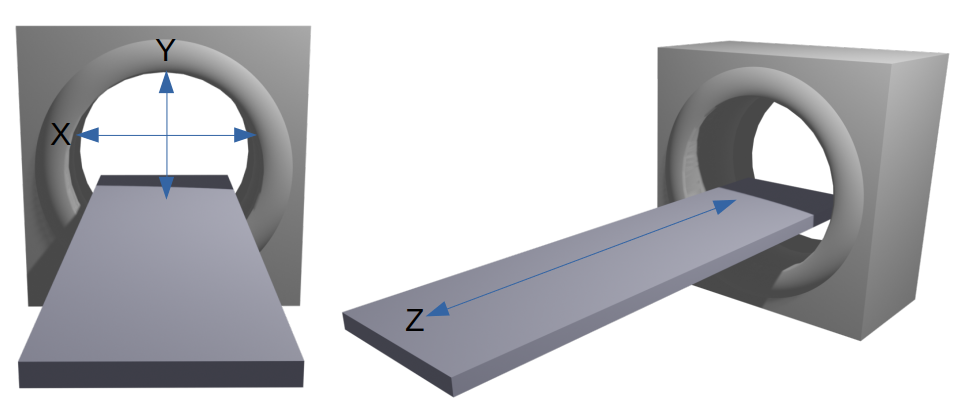}
\caption{Directional axes used in CT scanning}
\label{fig: ctaxes}
\medskip
\end{figure}

\hyperref[fig: ctaxes]{Figure} \ref{fig: ctaxes} illustrates the standard directional axes conventionally used in medical CT scanning to orient the subject of scanning on the bed. The $x$ axis extends across the width of the bed. The $y$ axis extends from ceiling to floor, and the $z$ axis runs along the length of the bed. The bed moves along the $z$ axis. An object that is oriented such that its longest dimension aligns along the $z$-axis will result in a better scan than an object with its shortest dimension aligned along the $z$-axis. This is an important consideration for scanning objects such as long bone fragments that tend to be longer than they are wide.

\begin{figure}[ht]
\centering
\captionsetup{margin=2cm, justification=centering}
\includegraphics[width=0.55\textwidth]{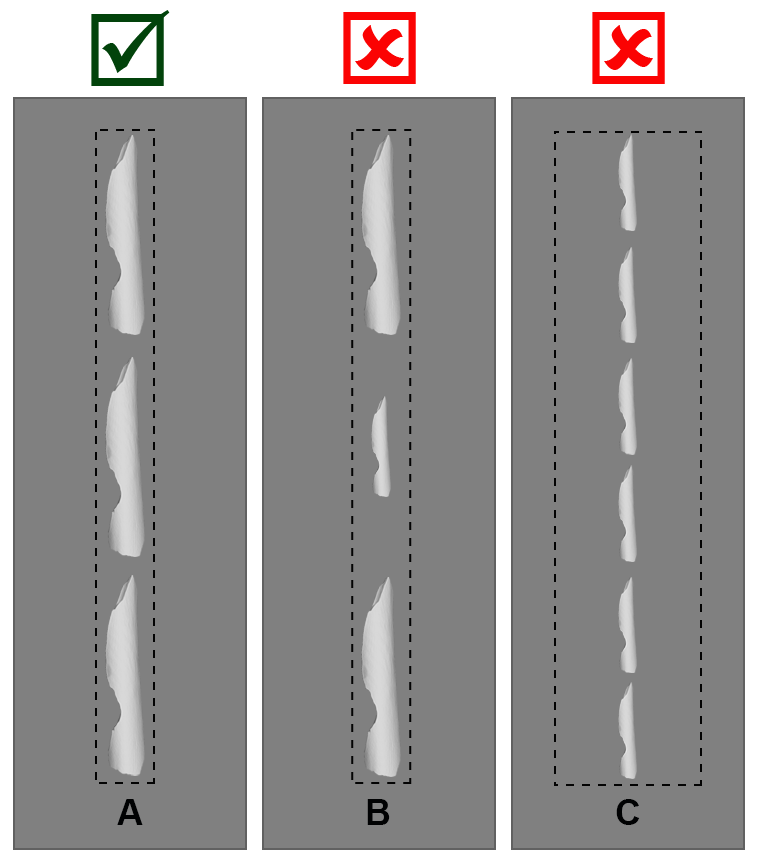}
\caption[Field of view]
{Field of view \medskip \par \small For the best resolution, the boundaries of the field of view should be as close to the target objects as possible (A). If specimens are disparate in size, the resolution of the smaller specimens will diminish (B). If the field of view is wide, this will also compromise resolution (C).}
\label{fig: field}
\medskip
\end{figure}

\subsection{Additional Considerations when CT scanning}

The Hounsfield Unit (HU) is a key concept of CT scanning that refers to radiographic density \parencite{scherf2013computed}. Materials are associated with specific Hounsfield units. Water sets the starting point at $0$ HU. Materials with lower radiodensity such as Fat have a negative Hounsefield unit ($-120$ to $-90$). Cancellous bone ranges from $300$ to $400$ HU and cortical bone ranges from $500$ to $1,900$ HU, therefore it is necessary to consider variation in the composition of the object to be scanned. This is especially important when CT imaging osteological materials and fossils coming from archaeological contexts. Bones that have fossilized may require different HU values than fresh bone \parencite{spoor2000using}. Additionally, if adhering matrix has a similar Hounsfield unit as the fossil, then they will be difficult to distinguish using CT. Conversely, in situations where the HU values are different, CT can be a way to ``remove'' adhering matrix without damaging the fossil \parencite{zollikofer1998computer, conroy1984noninvasive}. In fact, this is something to bear in mind when choosing the materials for creating scan packets --- one needs to ensure that the HU differs from that of the target object, whether it be adhering matrix or packaging material. Many material types can be captured using computed tomography and is not limited to bones or fossils. As an example, \textcite{goldner2022practical} established that micro-computed tomography (mCT) -- and thus medical CT -- can be used to scan stone tools and \textcite{van2005computed} offer an overview of objects scanned in archaeology (e.g. sarcophagi and bronze statues), soil science, the timber industry, industrial inspection, and aviation security. The key to applying our methods is to ensure that the packaging material can be separated based on a threshold value, i.e. the packaging material must be of a distinguishably different density than the object being scanned. 

An important parameter for the CT scanner is the field of view \parencite{miyata2020influence}. As the package size in which the fragments are placed increases, the field of view required by the scanner increases along with it. If this is due to an increase in the number of fragments in the packet, then the disparity in the size of each individual fragment and the overall field of view causes the quality of the scan to decrease. This is less of an issue if it is related to an increase in fragment size. By narrowing the field of view so that the scan is tightly focused around the line of bone fragments on the scanning bed, we can obtain higher resolution images compared to using a field of view that encompasses the whole width of the bed (see \hyperref[fig: field]{Figure} \ref{fig: field}). For this reason, we chose to create scan packets with similarly sized fragments, and, because the fragments were generally small, we chose to limit the number of fragments in each packet. Our CT scans have a resolution of 0.6mm between slices (along the direction of the scanning bed), and approximately 0.15mm resolution within slices with a narrow field of view. It is possible that with a much higher resolution CT scanner, multiple packets could be scanned in parallel while maintaining sufficient resolution which would substantially decrease overall time costs. 

Computational expenses are another important consideration and are largely centered on memory and processing power (e.g.~memory, storage, GPU, and speed). We have found that DICOM files require approximately 100MB of storage space per bone fragment, but of course this depends on resolution. Thus, the collection described in this paper takes roughly 250GB of storage space for the DICOM files on disk. After surfacing to create 3D triangulated surfaces for each object, the resulting meshes take on average 10MB per fragment. For our purposes, we purchased two 2TB external drives to transport files from the scanning facility so that we could surface them on our computers. The Batch Artifact Scanning Protocol does not require any visualization software packages as part of the post processing, which generally require some form of GPU. For example, Geomagic requires a GPU with a minimum of 2GB of memory, and in some cases 4GB, and Aviso requires a GPU with a minimum of 1GB. Running the surfacing step of the Batch Artifact Scanning Protocol involves purely CPU computations, and hence the post-processing can be performed on any computer without a GPU. On a high end laptop computer (MacBook Pro, 2 GHz Quad-Core Intel Core i5, 32 GB RAM) we were able to process each bone fragment in 4.375 seconds. Of course, faster computers and parallel processing can be used to accelerate the process, as needed.

\begin{figure}[ht]
\centering\captionsetup{margin=2cm, justification=centering}
\includegraphics[width=0.75\textwidth]{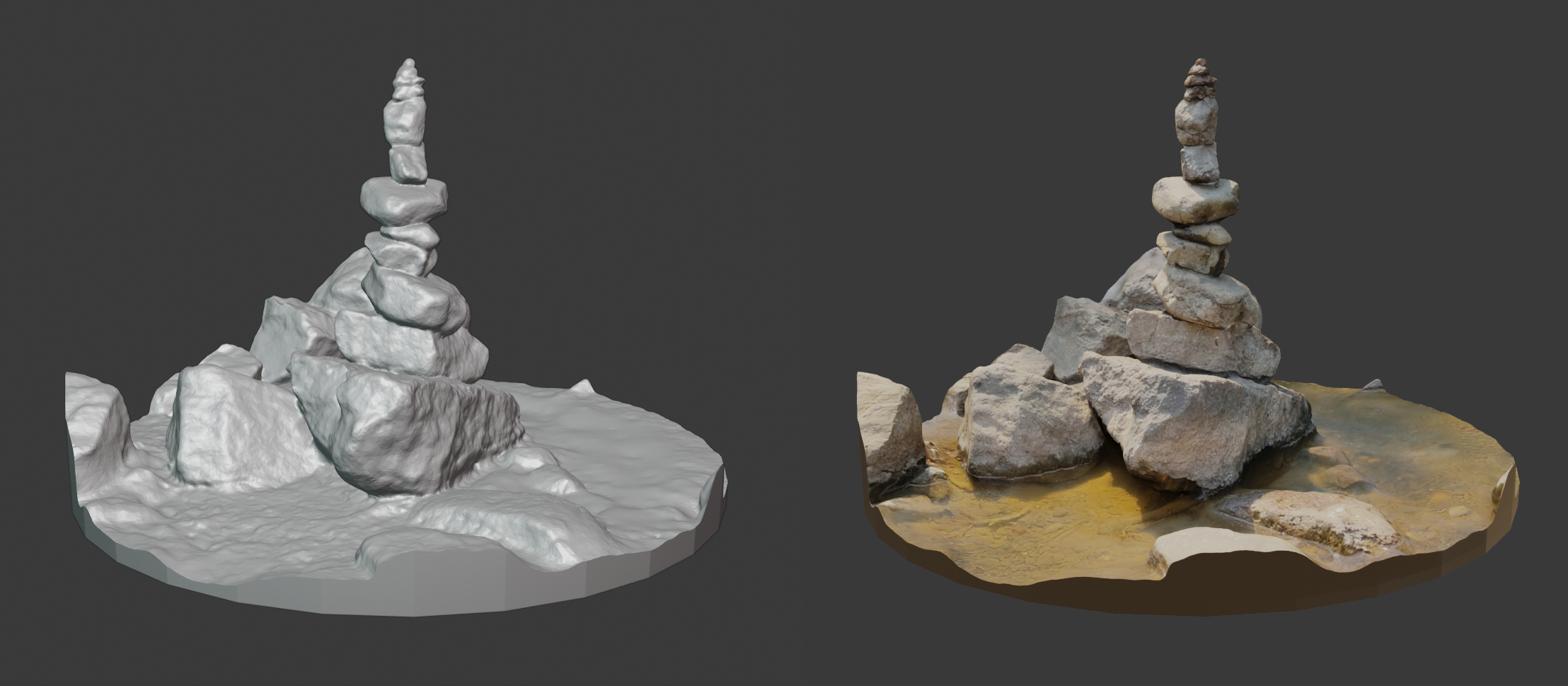}
\caption[Texture]
{Texture \medskip \par \small This a 3D mesh of a rock cairn at Gooseberry Falls. The image on the left is without texture. The image on the right has texture. Scanning and 3D model created by Dr. Samantha Porter.}
\label{fig: sam}
\medskip
\end{figure}

Beyond the logistics of scanning, it is necessary to consider what types of data need to be collected from the object. In particular, one must decide whether texture is required, determine the level of resolution required to answer the research question, and choose the parts of the object that need to be captured, e.g. whether or not this includes internal structures \parencite{bernardini20023d}.
In the simplest description, image texture refers to the perceived textures that are visible when looking at the object in real life (see \hyperref[fig: sam]{Figure} \ref{fig: sam}) and, in image processing, are defined by a series of texture units that describe a pixel (vertex or voxel) and its neighborhood \parencite{he1991texture}. Because our research focuses on the analysis of shape, we had no need to capture texture, making CT a viable option. The laser scanners and structured light scanners can oftentimes capture texture, however, this will increase processing times. The estimates provided in \hyperref[tab:time]{Table} \ref{tab:time} are based on fragments that were, in both cases, scanned without capturing texture. 

Resolution can be thought of as the level of detail present in the model; higher resolution offers more detail. A mesh is comprised of a certain number of points, oftentimes referred to as vertices, and the interpolated information in between those points. More points within in a given area increase the detail of the model. That said, higher resolution is not always necessary and adds time to computational processes when working with the model. If one were to model a flat plane, only three points would be necessary to uniquely specify it. However, as the object increases in complexity, more points are needed to capture that complexity. The resolution of the scanner is the limiting factor determining what can be expected for the resolution of the final 3D model. In our case, the CT scanner offered a resolution of 0.6mm between slices and approximately 0.15mm within slices (see Supplemental Information for more details).

\nc Because we needed more global features that did not require minute detail, a medical CT was sufficient. If a higher resolution is required then the post-processing methods presented could be applied to mCT scan data. We have code that can be applied to .DICOM and .tiff files. Though file types may change the applicability of our protocol, it would only require a very small change to the code to make this adjustment. 

An equally important consideration pertaining to resolution is scale. One can imagine zooming in on an image of the eastern coastline of Florida, as in \textcite{mandelbrot1975stochastic}. As one zooms in, the general outline between land and water will appear, then more curves along the shoreline will become visible, and ultimately one would be able to see the outline of individual grains of sand. If all that is needed is the general outline, then it would be computationally expensive to capture detail enough to see the grains of sand. Therefore, it is wise to consider the scale at which the research is begin conducted and the required level of detail. That said, the Batch Artifact Scanning Protocol described here can be effectively applied to images captured using a mCT with a flat scanning bed. Moreover, recent developments, outlined in \textcite[preprint]{o2024masse}, shows how we are currently expanding upon this work to adapt the protocol for use with rotational scanners, further broadening its applicability and utility across diverse mCT setups.   

Consideration of the structural features that need to be captured may dictate which scanning method makes most sense. Structured light scanners and photogrammetry can only capture the outside surface of the objects, i.e. that which can be seen by the naked eye. On the other hand, CT captures the internal geometry making it useful for scanning internal anatomy or encased objects. Furthermore, capturing deep crevices can be challenging using structured light scanning. Long bone shaft fragments can sometime come in the form of cylinders which are more easily captured using CT. Researchers who wish to study internal structures such as endocrania, trabecular bone, and foramina require CT scans and other approaches to 3D imaging \parencite[e.g.][]{brauer2004virtual, conroy1984noninvasive, conroy2000endocranial}.

\subsection{Adapting this Approach}

We described in detail how to effectively package specimens for safe transport and quick scanning, however, this particular approach is not entirely necessary. What is required is that the object be placed in the packet such that the individual bounding box for that specimen's 3D model will contain no other specimen and that each specimen and its location is accurately recorded in the \emph{.csv} file. Furthermore, the density of the packaging needs to be different than that of the specimens so that the CT scanner can differentiate the objects from the packaging. Packaging specimens together may not make sense for much larger objects. Packaging is recommended to limit set-up time at the scanning facility which has the potential of increasing the financial expense, especially if the charge is per hour. And, it is important to keep the objects safe during transport and handling at the imaging facility. 

While we have thus demonstrated how the Batch CT Scanning Method can be applied to rapidly scan large collections of bone fragments, or other archaeological materials, using medical CT and scans from mCT can also be post-processed in the same manner.


\section{Conclusion}
Here we have presented the Batch Artifact Scanning Protocol, a new method for rapidly scanning and automatically surfacing large collections. While demonstrated here with ungulate bone fragments, this approach is broadly applicable to any material scanned using CT or mCT technologies. Its potential is particularly significant in fields like zooarchaeology and taphonomy where collections can be quite large, oftentimes exceeding $10,000$ specimens. Additionally, the Batch Artifact Scanning Protocol can expedite the increasingly important push toward data sharing and the building of large online databases, thereby making powerful data analytical tools such as machine learning viable options for areas within archaeology (and more broadly anthropology) that have previously suffered from insufficient sample sizes. The Batch Artifact Scanning Protocol also has important implications for cultural heritage, education, and public-facing institutions such as museums. Collections can be scanned efficiently saving institutions time and money, while preservation of materials dramatically improves when researchers can use 3D models instead of handling the actual objects. 3D models can be used for educational purposes in formal and non-formal settings thus fostering interactive and other compelling connections with the broader public. \\


\begin{footnotesize}
\begin{noindent}
\textbf{Acknowledgments.} 
We would like to thank all who helped bring this project to fruition. The bone fragments used in scanning were sourced from Scott Salonek with the Elk Marketing Council and Christine Kvapil with Crescent Quality Meats. Bones were broken by hyenas at the Milwaukee County Zoo and Irvine Park Zoo in Chippewa Falls, Wisconsin and various math and anthropology student volunteers who broke bones using stone tools. Sevin Antley, Alexa Krahn, Monica Msechu, Fiona Statz, Emily Sponsel, Kameron Kopps, and Kyra Johnson helped clean, curate and prepare fragments for scanning. Thank you to Cassandra Koldenhoven and Todd Kes in the Department of Radiology at the Center for Magnetic Resonance Research (CMRR) for CT scanning the fragments. Pedro Angulo-Uma\~{n}a and Carter Chain worked on surfacing the CT scans. Matt Edling and the University of Minnesota’s Evolutionary Anthropology Labs provided support in coordinating sessions for bone breakage and guidance for curation. Abby Brown and the Anatomy Laboratory in the University of Minnesota’s College of Veterinary Medicine provided protocols and a facility to clean bones. Thank you to Samantha Porter who provided images and feedback on the manuscript. \\

\end{noindent}
\end{footnotesize} 

\begin{footnotesize}
\begin{noindent}

\noindent
\textbf{Competing Interests:} The authors declare none.\\

\noindent
\textbf{Funding Information.} We would like to thank the National Science Foundation NSF Grant DMS-1816917, NSF SBE SPRF 2204135, and the University of Minnesota's Department of Anthropology for funding this research. Calder was supported by NSF grants DMS:1944925 and MoDL+ CCF:2212318, the Alfred P. Sloan foundation, the McKnight foundation, and an Albert and Dorothy Marden Professorship. \\

\noindent
\textbf{Data Availability.} Source code can be found at  \url{https://github.com/jwcalder/CT-Surfacing}. 

\end{noindent}
\end{footnotesize}


\printbibliography

\newpage

\includepdf[pages=-]{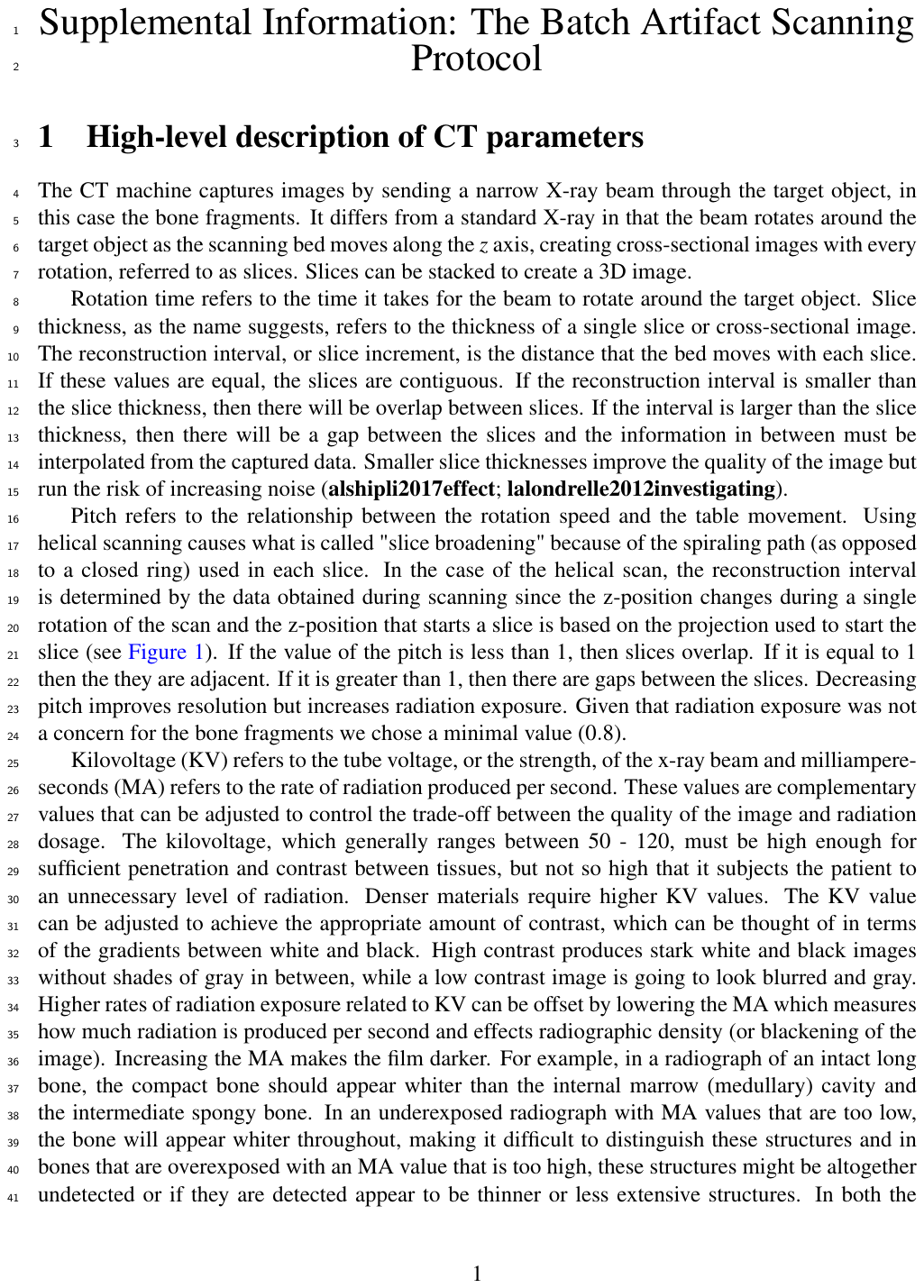}
\setcounter{section}{0} 
\setcounter{figure}{0} 

\end{document}